# PrivLogit: Efficient Privacy-preserving Logistic Regression by Tailoring Numerical Optimizers

Wei Xie
Vanderbilt University
Nashville, TN
wei.xie@vanderbilt.edu

Yang Wang
University of Virginia
Charlottesville, VA
yw3xs@virginia.edu

Steven M. Boker
University of Virginia
Charlottesville, VA
boker@virginia.edu

Donald E. Brown
University of Virginia
Charlottesville, VA
deb@virginia.edu

## Abstract

Safeguarding privacy in machine learning is highly desirable, especially in collaborative studies across many organizations. Privacy-preserving distributed machine learning (based on cryptography) is popular to solve the problem. However, existing cryptographic protocols still incur excess computational overhead. Here, we make a novel observation that this is partially due to naive adoption of mainstream numerical optimization (e.g., Newton method) and failing to tailor for secure computing. This work presents a contrasting perspective: customizing numerical optimization specifically for secure settings. We propose a seemingly less-favorable optimization method that can in fact significantly accelerate privacy-preserving logistic regression. Leveraging this new method, we propose two new secure protocols for conducting logistic regression in a privacy-preserving and distributed manner. Extensive theoretical and empirical evaluations prove the competitive performance of our two secure proposals while without compromising accuracy or privacy: with speedup up to 2.3x and 8.1x, respectively, over state-of-the-art; and even faster as data scales up. Such drastic speedup is on top of and in addition to performance improvements from existing (and future) state-of-the-art cryptography. Our work provides a new way towards efficient and practical privacy-preserving logistic regression for large-scale studies which are common for modern science.

## 1 Introduction

Data integration and joint analytics (i.e., machine learning) in a distributed network of independent databases (belonging to different organizations) is widely used in various disciplines, such as in data management [Zhang and Zhao, 2005, Verykios et al., 2004, Aggarwal and Philip, 2008], biomedical and social sciences [McCarty et al., 2011, Zheng et al., 2017] where data are distributed in nature and single databases only have limited samples sizes. The goal of such multi-organization joint analytics is to accumulate large sample sizes across databases/organizations and reach more powerful machine learning conclusions.

Such practice of joint analytics across distributed multi-organization databases, however, is often hampered by serious privacy issues as these studies typically involve human subject data (e.g., electronic medical records, human genome) that are sensitive and strictly protected by various privacy laws and regulations [Office for Civil Rights, 2002, Hudson et al., 2008, Daries et al., 2014]. Meanwhile, many participating orga-nizations are also reluctant to reveal their data content to external entities (due to concerns around privacy and business secrets), even though they still want to contribute to collaborative studies. This is increasingly common in many disciplines concerning data management, such as healthcare and biomedicine [Li et al., 2015, Wu et al., 2012b], business, finance, etc.

More formally, we are interested in the following common scenario (traditionally known as *privacy-preserving distributed data mining* in data management and machine learning [Verykios et al., 2004, Zhang and Zhao, 2005, Aggarwal and Philip, 2008]): multiple independent organizations (e.g., different institutions, medical centers) want to conduct joint analytics (such as logistic regression). They each possess their respective *private* data of a sub-population (e.g., electronic medical records, human genomes, or psychology surveys), but are not willing or permitted to disclose the data beyond their respective organizations due to privacy and proprietary reasons. We focus on the *horizontally partitioned* setting [Aggarwal and Philip, 2008], where each independent database contains only a sub-population (i.e., some rows). In such a collaborative study across distributed databases, potential adversaries include: distrustful aggregation center (e.g., due to breached servers or malicious employees), distrustful member organizations (due to curiosity about other organizations' secrets or business competition), and external curious people or hackers. The adversary's goal is to learn privacy-sensitive information of individual data records or organizations by peeking into raw and summary-level data. The challenge here is on how to support such a collaborative study while preserving privacy, especially when it is difficult or economically impractical to find a fully entrusted central authority to directly aggregate ("union") all databases.

Privacy-preserving distributed data mining leveraging cryptography (secure multi-party computation in particular) and distributed computing is a classical and reviving solution for tackling the challenge [Aggarwal and Philip, 2008]. Numerous efforts have attempted to support data mining in distributed databases without disclosing raw and intermediate data [Aggarwal and Philip, 2008, Nardi et al., 2012, Nikolaenko et al., 2013, Xie et al., 2014, Boker et al., 2015, Li et al., 2016]. Among these, significant attention is devoted to logistic regression [Wolfson et al., 2010, El Emam et al., 2013, Nardi et al., 2012, Li et al., 2016, Aono et al., 2016], one of the most popular statistical methods.

Despite encouraging progress, very few proposals have seen wide adoption in real world for privacy-preserving logistic regression. The main reason is still the excess computational overhead of cryptographic methods, despite recent improvements in cryptography alone (represented and partially summarized by [Liu et al., 2015]). While it is generally expected for secure computation to be slower than non-secure counterparts, we also make a *novel and surprising* observation: much of the computational overhead indeed traces back to the *sub-optimal* technical decisions made by humans experts (e.g., authors of cryptographic solutions) and could have been avoided. For instance, nearly all existing secure protocols [El Emam et al., 2013, Li et al., 2016] *directly apply* mainstream (distributed) model estimation algorithms (e.g., the popular Newton method for logistic regression [Harrell, 2015]), failing to account for secure computing-specific characteristics and thus missing valuable opportunities for performance improvement. Our work here is motivated to leverage our novel observation and correct for a *wide-spread suboptimal practice* in privacy-preserving data mining.

In this work, we propose a different approach to making privacy-preserving logistic regression much more efficient and practical: Instead of following common practice of taking "off-the-shelf" numerical optimization algorithms (e.g., Newton method) and focusing on accelerating the underlying cryptography alone (pervasive among nearly all related secure protocols [Aggarwal and Philip, 2008]), we propose to *tailor numerical optimization specially for cryptographic computing in general (but not tied to any specific cryptographic schemes or implementations), which then can be built upon whatever state-of-the-art cryptography available now and in future*. Our new approach significantly accelerates overall computation in addition to enjoying latest advancement from state-of-the-art cryptography, while guaranteeing privacy and result accuracy.

In our proposed new numerical optimizer (termed PrivLogit), we derive a constant approximation for the second-order curvature information (i.e., Hessian) in the Newton method for logistic regression. This adapted optimizer seems counter-intuitive and "unfavorable" due to its elongated convergence and increased network interactions, but surprisingly turns out to be highly competitive in overall performance. Following PrivLogit, we propose and evaluate two highly-efficient cryptographic protocols for privacy-preserving distributed logistic regression, i.e., PrivLogit-Hessian and PrivLogit-Local.

Extensive theoretical and experimental evaluations show that our method is significantly more efficient and practical, make secure data management closer towards real-world deployment.

**Contributions** Our contributions are as follows:

- We make a novel observation about a generic performance bottleneck in privacy-preserving logistic regression, i.e., directly building on de facto optimization algorithms originally designed for *non-cryptography* settings. This interesting observation has long been neglected by the domain.

- Based on above observation, we propose a counter-intuitive but surprisingly much better model estimation method (i.e., PrivLogit) for privacy-preserving logistic regression. This provides a entirely different perspective for privacy-preserving distributed logistic regression, by showing that tailoring numerical optimizers for secure computing can lead to *significant unexpected* performance gains.

- We propose two highly-efficient secure protocols (i.e., PrivLogit-Hessian and PrivLogit-Local) for privacy-preserving logistic regression, which significantly outperforms state-of-the-art alternatives based on the same latest cryptographic schemes and tools.

- We provide detailed theoretical proof and analysis on our proposals.

- We extensively evaluate our proposals on various simulated and real-world studies of very large scale, many of which are significantly larger than previously reported in the literature. As a side result, we also present and evaluate the first comprehensive secure implementation/protocol of Newton method based on latest cryptography (as our baseline).

**Outlines** We first provide background on logistic regression and model estimation methods in Section 2. In Sections 3 and 4, we describe our improved optimization method PrivLogit, and two secure implementations. In Section 5, we elaborate on theoretical details regarding model convergence proof, security analysis, and computational complexity of our proposals. This is followed by experimental results in Section 6. In Section 7, we survey related works. We discuss and conclude in Section 8.

## 2 Preliminaries

This manuscript roughly follows the notations in Table 1.

### 2.1 Logistic Regression

This work concerns conducting logistic regression in a multi-organization (distributed) environment. Logistic regression is a probabilistic model that is often used for binary (i.e., categorical) classification [Harrell, 2015]. It is among the most utilized statistical models in practice, with wide adoption in biomedicine [Lowrie and Lew, 1990], genetics [Lewis and Knight, 2012], online advertising [McMahan et al., 2013], etc. Briefly, the logistic regression is defined as:

$$p(y = 1|\mathbf{x}; \boldsymbol{\beta}) = \frac{1}{1 + e^{-\boldsymbol{\beta}^T \mathbf{x}}}, \quad (1)$$

where $p(.)$ denotes the probability of the binary response variable *y* equal to 1 (i.e., "case" or "success" in practice), $\mathbf{x}$ is the $p$-dimensional covariates for a specific data record, and $\boldsymbol{\beta}$ is the $p$-dimensional regression coefficients we want to estimate.

Table 1: Notations.

| Notations | Remarks |
|---|---|
| $\mathbf{X} \in \mathbb{R}^{n \times p}$ | Regression covariates: $n$ samples, $p$ features |
| $\mathbf{y} \in \mathbb{R}^n$ | Regression response vector: $n$ samples |
| $\boldsymbol{\beta} \in \mathbb{R}^p$ | Regression coefficients |
| $\mathbf{H}, \tilde{\mathbf{H}} \in \mathbb{R}^{p \times p}$ | Hessian, approximate Hessian matrices |
| $\mathbf{g} \in \mathbb{R}^p$ | Gradient |
| $\lambda \in \mathbb{R}$ | Regression regularization parameter |
| $l_2(\boldsymbol{\beta})$ | Log-likelihood (objective) |
| $Enc(data)$ | Encryption of $data$ |
| $\oplus, \ominus, \otimes, \oslash$ | Secure arithmetics for $+, -, \times, \div$ |
| $E_{sqrt}(data)$ | Secure square root of $data$ |

In practice, regularization is often applied to the model estimation process to aid feature selection and prevent overfitting by penalizing extreme parameters [Nigam, 1999]. Here we consider the popular $\ell_2$-regularization for logistic regression [Nigam, 1999] to make our work generically applicable. The standard logistic regression can be derived by simply setting the regularization to 0. The $\ell_2$-regularized logistic regression imposes an additional regularization term, $-\frac{\lambda}{2}\boldsymbol{\beta}^T\boldsymbol{\beta}$, to the optimization objective during model estimation. For a dataset $(\mathbf{X} \in \mathbb{R}^{n \times p}, \mathbf{y} \in \mathbb{R}^n) = [(\mathbf{x_1}, y_1), ..., (\mathbf{x_n}, y_n)]$ with $n$ independent samples and $p$ features, the log-likelihood (i.e., optimization objective) of $\ell_2$-regularized logistic regression is:

$$l_2(\boldsymbol{\beta}) = \sum_{i=1}^{n}[y_i(\boldsymbol{\beta}^T \mathbf{x_i}) - \log(1 + e^{\boldsymbol{\beta}^T \mathbf{x_i}})] - \frac{\lambda}{2}\boldsymbol{\beta}^T\boldsymbol{\beta}, \quad (2)$$

where $\lambda$ is the predefined penalty parameter to tune the regularization.

## 2.2 Distributed Newton Method

When fitting a (regularized) logistic regression, the goal is to estimate the coefficients $\boldsymbol{\beta}$ from existing training data $(\mathbf{X}, \mathbf{y})$. Since logistic regression does not have a closed form, model estimation is often accomplished by (iterative) numerical optimization over the objective $l_2(\boldsymbol{\beta})$. The *de facto* approach for estimating the (regularized) logistic regression coefficient $\boldsymbol{\beta}$ (Equation 1) is the Newton method [Harrell, 2015], which is widely implemented in most statistical software and underlying nearly all existing protocols for privacy-preserving logistic regression (e.g., [Wolfson et al., 2010, El Emam et al., 2013, Li et al., 2016]). Newton method iteratively approaches the optimal coefficients, and for each iteration, the coefficient estimates are updated by:
$$\boldsymbol{\beta}^{(t+1)} = \boldsymbol{\beta}^{(t)} - \mathbf{H}^{-1}(\boldsymbol{\beta}^{(t)}) \, \mathbf{g}(\boldsymbol{\beta}^{(t)}) \,, \tag{3}$$

where $\mathbf{H}(\boldsymbol{\beta}^{(t)})$ and $\mathbf{g}(\boldsymbol{\beta}^{(t)})$ denote the Hessian and gradient of the objective $l_2(\boldsymbol{\beta})$ (Equation 2) evaluated at the current $\boldsymbol{\beta}^{(t)}$ coefficient estimate. The superscripts $(t), (t+1)$ denote the $t^{\text{th}}, (t+1)^{\text{th}}$ iterations, respectively. This updating process iterates until model convergence.

Based on Equation 2, the gradient and Hessian for $\ell_2$-regularized logistic regression can be computed as follows (setting $\lambda = 0$ will skip regularization and yield the standard logistic regression):

$$\mathbf{g}(\boldsymbol{\beta}) = \nabla_{\boldsymbol{\beta}} l_2(\boldsymbol{\beta}) = \mathbf{X}^T(\mathbf{y} - \mathbf{p}) - \lambda \boldsymbol{\beta} = \sum_{j=1}^{S} \mathbf{g_j}(\boldsymbol{\beta}) - \lambda \boldsymbol{\beta} \,, \tag{4}$$

$$\mathbf{H}(\boldsymbol{\beta}) = \frac{d^2 l_2(\boldsymbol{\beta})}{d\boldsymbol{\beta} d\boldsymbol{\beta}^T} = -\mathbf{X}^T \mathbf{A} \mathbf{X} - \lambda \mathbf{I} = \sum_{j=1}^{S} \mathbf{H_j}(\boldsymbol{\beta}) - \lambda \mathbf{I} \,, \tag{5}$$

where $\mathbf{X}$ represents covariates of $n$ samples and $p$ features; $\mathbf{y}$ denotes the response vector of $n$ data records; $\mathbf{p} \in \mathbb{R}^n$ is the vector of logistic regression probabilities for $n$ records; $\mathbf{A} \in \mathbb{R}^{n \times n}$ is a diagonal matrix with elements defined as $a_{i,i} = p_i(1 - p_i)$; and $\mathbf{g_j}(\boldsymbol{\beta})$ and $\mathbf{H_j}(\boldsymbol{\beta})$ are the per-organization gradient and Hessian, respectively, that will be introduced afterwards in the distributed version; $S$ is the total number of organizations contributing data to the collaborative study.

As is also manifested in the last equalities of Equations 4 and 5, the computation of both $\mathbf{g}(\boldsymbol{\beta})$ and $\mathbf{H}(\boldsymbol{\beta})$ can be decomposed per participating organizations (who can freely access their respective private data such as $\mathbf{X}_j, \mathbf{y}_j$), and thus need not invoke expensive cryptographic computation (except for the final summation across organizations).

## 3 PrivLogit: a Tailored Optimizer for Fast Privacy-preserving Logistic Regression

Here, we first analyze the problems with mainstream secure Newton method that is underlying nearly all existing solutions for privacy-preserving logistic regression. This motivates us to customize a better optimization method (PrivLogit) specially for secure computing (but not tied to any particular cryptographic scheme or implementation). We later analyze the attractive properties of PrivLogit, which seem obscure at first sight.

### 3.1 Limitations of Newton Method.

To estimate regression coefficients via the aforementioned Newton method (Equations 3), the evaluation and inversion of the Hessian have to be repeated for every iteration until model convergence. These two operations can be prohibitively expensive in computation and network communication especially when implemented using cryptography.

For (distributed) Newton method in general (e.g., non-secure applications), it has been well acknowledged that the evaluation and inversion of the Hessian matrix are the overall computational bottleneck due to large data sizes, inherent complexity and repetitive nature of these operations [Liu and Nocedal, 1989]. This in fact has motivated numerous improved optimizers in machine learning and optimization [Liu and Nocedal, 1989]. *Unfortunately, most such newer optimizers do not seem amenable to efficient and data-agnostic secure implementation and thus are not used by the cryptography and privacy community).*

In data security and privacy research, the issue of expensive Newton method is exacerbated as secure inversion of Hessian matrix requires complex operations (e.g., secure division and square root) which have to resort to expensive primitives and approximations from secure multi-party computation [Nardi et al., 2012, El Emam et al., 2013]. As a result, almost all existing secure logistic regression proposals have to compromise privacy protection or result accuracy to increase performance (e.g., to selectively reveal intermediate data/computations [Li et al., 2016] or to use approximations [Nardi et al., 2012, Aono et al., 2016]).

### 3.2 PrivLogit for Privacy-preserving Logistic Regression.

We are motivated to design a tailored optimizer for secure computing by addressing the aforementioned limitations of Newton method. Our proposal is based on the classical work on quadratic function approximation [Böhning and Lindsay, 1988] (non-secure applications) and with new theoretical analysis. In brief, we propose to use one carefully-chosen constant matrix as a surrogate for the exact Hessian matrices (Equation 5) across all iterations. Specifically, the following approximate Hessian (denoted $\tilde{\mathbf{H}}$) is proposed:

$$\tilde{\mathbf{H}} = -\frac{1}{4}\mathbf{X}^T\mathbf{X} - \lambda \mathbf{I} \,, \tag{6}$$

here $\tilde{\mathbf{H}}$ is a tight lower bound because for all $p_i \in [0,1]$ (the probability in logistic regression), we have that: $\max\{a_{i,i} = p_i(1-p_i)\} = \frac{1}{4}$ (where $a_{i,i}$ denotes elements of the diagonal matrix $\mathbf{A}$ defined in Equation 5 for Hessian). *We highlight that this approximation guarantees exact model convergence and result accuracy (with theoretical and empirical proof later in Sections 5.1 and 6.2).*

The calculation of approximate Hessian $\tilde{\mathbf{H}}$ can be decomposed per-organization (horizontally partitioned) and computed in a distributed manner among many organizations:

$$\tilde{\mathbf{H}} = -\frac{1}{4}\sum_{j=1}^{S}\mathbf{X_j}^T\mathbf{X_j} - \lambda\mathbf{I} = \sum_{j=1}^{S}\tilde{\mathbf{H}}_j - \lambda\mathbf{I} \tag{7}$$

where $\mathbf{X}_j$ is the (privacy-sensitive) raw data stored locally at Organization $j$, $S$ is the total number of organizations contributing data, and $\tilde{\mathbf{H}}_j = -\frac{1}{4}\mathbf{X}_j^T\mathbf{X}_j$ denotes the approximate Hessian for Organization $j$.

Substituting this approximate Hessian into the Newton method (Equation 3), along with the distributed evaluation of gradient (Equation 4), the iterative updating formula for our new optimizer (denoted as PrivLogit) follows:

$$\boldsymbol{\beta}^{(t+1)} = \boldsymbol{\beta}^{(t)} - [\sum_{j=1}^{S}\tilde{\mathbf{H}}_j - \lambda\mathbf{I}]^{-1}[\sum_{j=1}^{S}\mathbf{g}_j(\boldsymbol{\beta}^{(t)}) - \lambda\boldsymbol{\beta}^{(t)}] \tag{8}$$

The above iterative process continues until model convergence. Convergence can be measured by the relative change of log-likelihood and compared against a predefined threshold (e.g., $10^{-6}$): $\frac{|l_2^{(t+1)} - l_2^{(t)}|}{|l_2^{(t)}|} < 10^{-6}$ , where $l_2^{(t+1)}, l_2^{(t)}$ correspond to the log-likelihood of logistic regression for Iterations $(t+1), (t)$, respectively. Note that based on Equation 2, the log-likelihood can also be decomposed per-organization $j$ (whose share is denoted $l_{sj}$):

$$l_2(\boldsymbol{\beta}) = \sum_{j=1}^{S} l_{sj} - \frac{\lambda}{2}\boldsymbol{\beta}^T\boldsymbol{\beta} \tag{9}$$

### 3.3 Attractive Properties of PrivLogit.

Our new PrivLogit numerical optimizer enables a few attractive properties, which seem highly promising for efficient privacy-preserving logistic regression.

#### 3.3.1 Asymmetric Computational Complexity in Secure Settings

The PrivLogit adaption comes at the cost of more iterations required for convergence (and also increased local-organization computation), which seems counter-intuitive and less favorable as more iterations mean slower convergence. However, this view fails to consider computational cost as a whole and the different computational characteristics of distributed model estimation with and without cryptographic protections. In secure implementations, the local computation at each organization is essentially "free" because organizations have full control of their respective private data and (extremely) fast non-secure computations are applicable; but secure computation at the aggregation center is usually orders of magnitudes slower than non-secure counterparts (due to expensive cryptographic protections against an adversarial center). This implies that eliminating complexity of center-based secure computation (current bottleneck) can potentially lead to significant speedup (as is the case in PrivLogit). Our method has very cheap per-iteration cost, making it competitive in overall performance.

#### 3.3.2 Constant Hessian

Our proposed Hessian approximation stays constant and independent of the varying $\beta^{(t)}$'s coefficients across all iterations. This indicates that it only needs to be evaluated and inverted *once* during preprocessing and can then be reused across all iterations, leading to dramatic reduction in computation compared with traditional Newton method.

#### 3.3.3 Decomposition of Computation

The new optimizer allows for easy decomposition the computation among participating organizations, which can be leveraged to achieve significant speedup. For instance, the approximate Hessian can be computed in a distributed manner via a series of aggregations, as demonstrated in Equation 7. So is the gradient (Equation 4).

In addition, further reduction in computation is possible after the approximate Hessian is securely inverted and properly protected. As will be introduced later in our second implementation PrivLogit-Local (Section 4.2), partial Newton update direction can be computed locally by each local nodes (who has privacy-free access to their respective private data and thus local gradient need not be encrypted). The center only needs to securely aggregate these local Newton steps, which is highly efficient. *This performance improvement is unique to PrivLogit and not possible for Newton method due to varying Hessian's.*

#### 3.3.4 Guaranteed Model Quality

Despite the approximation to Hessian, the PrivLogit optimizer is guaranteed to converge to accurate model estimates, i.e., perfect accuracy (Sections 5.1 and 6.2).

# 4 Cryptographic Implementations of PrivLogit

Based on our new PrivLogit optimization method, we propose two cryptographic protocols for privacy-preserving distributed logistic regression. The first is called PrivLogit-Hessian which is a straightforward cryptographic implementation of PrivLogit. Our second proposal, called PrivLogit-Local, provides even more speedup by offsetting some expensive matrix operations to local organizations and taking advantage of their fast and privacy-free computing power.

### 4.0.1 Distributed System Architecture

Both our secure protocols adopt the distributed architecture consisting of local Nodes (organizations) and an aggregation Center (semi-honest), as illustrated in Figure 1 (common for privacy-preserving distributed data mining [Aggarwal and Philip, 2008]). In brief, participating organizations (i.e., distributed Nodes) are responsible for protecting their respective data and only generating (safe) summary-level data, which would be encrypted and securely consumed by the Center for model estimation. In a strongly protected system such as ours, *all data and computations at the Center are encrypted and not visible even to the Center itself*. The role of Center is typically played by two or more mutually independent semi-trusted authorities (denoted as different Servers in Figure 1), as is common for secure multi-party computation applications [Aggarwal and Philip, 2008, Nikolaenko et al., 2013, Xie et al., 2014, Li et al., 2016]. In practice, such as in biomedical or social sciences, Center(s) can be the coordinating center (of a consortium, federation or association) in addition to a third-party authority (e.g., audit organizations). We acknowledge there can be alternative architectures, such as Center with 3-parties [Kamm and Willemson, 2013] or even more, or fully decentralized. However, since this is not the focus of our work and state-of-the-art are still mostly using two-party (Center) architectures, we thus leave it as future work.

### 4.0.2 Choice of Cryptographic Schemes

Our proposals are agnostic of specific choices of cryptographic schemes and most existing or future cryptographic schemes/primitives can be utilized. This is because, as discussed in Section 3.3, our performance advantage comes from leveraging the *drastic computational complexity asymmetry* between the untrusted Center (whose computation is *orders of magnitude slower* [Liu et al., 2015] due to cryptography) and distributed Nodes (whose computation is often *privacy-free and extremely fast*). *This asymmetry is likely to exist for the foreseeable future regardless of improvements in cryptography alone. In fact, since PrivLogit has much simpler main computations (Equations 8 and 7) than Newton method (Equations 3 and 5), our proposals are more widely amenable to a variety of cryptographic schemes.* Since the focus of our work is not on specific cryptographic protocols and due to space constraint, we directly adopt state-of-the-art cryptographic schemes (building blocks) and skip trivial cryptographic details that are common knowledge in privacy-preserving (distributed) data mining [Aggarwal and Philip, 2008, Nikolaenko et al., 2013, Li et al., 2016, Xie et al., 2014].

Our current secure implementation builds on a hybrid of two widely-used cryptographic schemes: Yao's garbled circuit [Yao, 1982] (mainly for Type 2 computations between independent Center servers as depicted in Figure 1) and Paillier encryption [Paillier, 1999] (mainly for Type 1 computations between local Nodes and the Center as depicted in Figure 1). Local-node computations are mostly privacy-free. *The hybrid of garbled circuit and Paillier encryption (and the conversion between each other) is very efficient and is state-of-the-art cryptographic protocol for various closely related tasks, including privacy-preserving* logistic regression and other machine learning models [Aggarwal and Philip, 2008, El Emam et al., 2013, Nikolaenko et al., 2013, Xie et al., 2014].

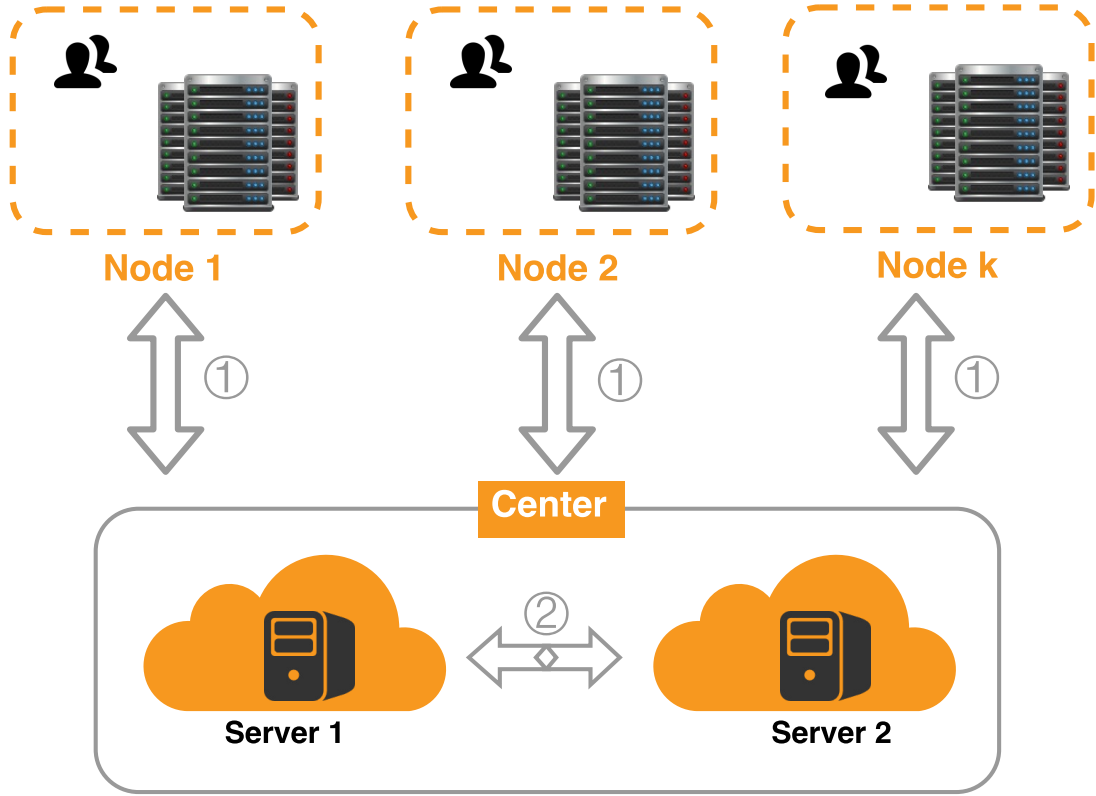

Figure 1: Distributed architecture for privacy-preserving logistic regression. Two main types of computations are involved between: 1) local Nodes and the Center; 2) different Servers/authorities at the Center.

### 4.0.3 Secure Notations

For simplicity, we use intuitive symbols to denote a few common secure mathematical arithmetics. Each of these operations take encrypted operands as inputs, and securely compute without decryption to output an encrypted result. As before, encrypted data are represented as $Enc(.)$. And we denote secure addition, subtraction, multiplication, and division as: $\oplus, \ominus, \otimes, \oslash$, respectively (secure square root $E_{sqrt}(.)$ is also used in matrix inversion whose details are omitted in main text).

Across all secure Algorithms 1, 2, and 3 introduced below, *we flag computations by their location of occurrence* in accordance with the distributed architecture in Figure 1. This also explains what cryptographic schemes are underlying what computations, because as mentioned earlier in Section 4, Center-based computations are primarily implemented using garbled circuit, and local node-computations are privacy-free, and the information exchange between Center and local Nodes is implemented using Paillier encryption.

## 4.1 PrivLogit-Hessian: Secure Distributed PrivLogit

As presented in Algorithm 1, PrivLogit-Hessian is our straightforward secure and distributed implementation of our PrivLogit optimizer. This secure protocol consists of two phases of computation: a one-time setup phase of securely approximating and inverting the Hessian, and a repeated (iterative) secure model estimation phase.

---

**Algorithm 1** PrivLogit-Hessian: Fast and Secure Logistic Regression.

---

**Input:** Random initial $\boldsymbol{\beta}^{(0)}$; Regularization parameter $\lambda$
**Output:** Globally fit coefficient estimate $\boldsymbol{\beta}$
   *[At local organizations and Center]* :
1: Securely approximate and Cholesky-decompose (negated) Hessian: $Enc(\mathbf{L}) = SetupOnce()$ (where $Enc(\mathbf{L}\mathbf{L}^T) = Enc(-\tilde{\mathbf{H}})$) (Algorithm 2)

2: **while** regression model not converged **do**
     *[At local organizations]* :
3:    **for** each organization j = 1 **to** S **do**
       *[At local Organization j]* :
4:       Compute local gradient $\mathbf{g}_j$ and encrypt via Paillier (Equation 4)
5:       Compute local log-likelihood $l_{sj}$ and encrypt via Paillier (Equation 9)
6:       Securely transmit encryptions $Enc(\mathbf{g}_j), Enc(l_{sj})$ to Center
7:    **end for**

     *[At Center]* :
8:    Securely aggregate gradients across organizations: $Enc(\mathbf{g}) = Enc(\mathbf{g}_1) \oplus ... \oplus Enc(\mathbf{g}_j) \oplus ... \oplus Enc(\mathbf{g}_S) \ominus Enc(\lambda \boldsymbol{\beta}^{(t)})$ (Equation 4)
9:    Secure back-substitution: $Enc(\tilde{\mathbf{H}}^{-1}\mathbf{g}) \leftarrow Enc(\mathbf{L})$, $Enc(\mathbf{g})$
10:   Securely update coefficient estimates via PrivLogit: $\boldsymbol{\beta}^{(t+1)} \leftarrow \boldsymbol{\beta}^{(t)}$ (Equation 8)
11:   Securely aggregate log-likelihood across organizations: $Enc(l_2) = Enc(l_{s1}) \oplus ... \oplus Enc(l_{sj}) \oplus ... \oplus Enc(l_{sS}) \ominus Enc(\frac{\lambda}{2}[\boldsymbol{\beta}^{(t)}]^T \boldsymbol{\beta}^{(t)})$ (Equation 9)
12:   Securely check model convergence using secure comparison (Section 3.2)
13:   Securely disseminate new coefficient estimates to each local organizations: $Enc(\boldsymbol{\beta}^{(t+1)})$
14: **end while**
15: **return** $\boldsymbol{\beta}^{(t)}$ (last converged estimate)

---

The first phase (Step 1 in Algorithm 1 or *SetupOnce()* function in Algorithm 2) focuses on securely approximating and inverting Hessian. Specifically, based on Algorithm 2, each local organizations compute their local Hessian approximation $\tilde{\mathbf{H}}_j$ (based on covariance matrix $\mathbf{X}_j^T \mathbf{X}_j$) and encrypt it before sharing with the Center (Steps 1 to 4 in Algorithm 2; our current implementation uses Paillier encryption). The Center securely aggregates these encrypted per-organization Hessians (and the regularization term as necessary), yielding an encrypted global Hessian approximation $Enc(\tilde{\mathbf{H}})$ (Step 5 in Algorithm 2 and Equation 7; encryption is still based on Paillier). Later, the Center needs to securely invert the Hessian, which is typically achieved by secure Cholesky decomposition on the protected (negated) Hessian and obtains its encrypted "inversion" (the encrypted Cholesky triangular matrix $Enc(\mathbf{L})$ to be precise), such that $Enc(\mathbf{L}\mathbf{L}^T) = Enc(-\tilde{\mathbf{H}})$. Since Cholesky decomposition is a common textbook algorithm and has been implemented before [Nikolaenko et al., 2013] using the same cryptography as ours (garbled circuit),

---

**Algorithm 2** *SetupOnce()* for securely approximating and inverting Hessian.
---
**Input:** Local organizations with their respective data
**Output:** Encrypted triangular matrix $Enc(\mathbf{L})$ from Cholesky decomposition (where $Enc(\mathbf{LL}^T) = Enc(-\tilde{\mathbf{H}})$)
   *[At local organizations]*:
1: **for** each organization $j$ = 1 **to** S **do**
2:    Approximate local Hessian $\tilde{\mathbf{H}}_j$ (Equation 6)
3:    Encrypt and securely transmit $Enc(\tilde{\mathbf{H}}_j)$ to Center
4: **end for**

   *[At Center]*:
5: Securely aggregate Hessians across organizations: $Enc(\tilde{\mathbf{H}}) = Enc(\tilde{\mathbf{H}}_1) \oplus ... \oplus Enc(\tilde{\mathbf{H}}_j) \oplus ... \oplus Enc(\tilde{\mathbf{H}}_S) \ominus Enc(\lambda \mathbf{I})$ (Equation 6)
6: Secure Cholesky decomposition to obtain: $Enc(\mathbf{L})$ (where $Enc(\mathbf{LL}^T) = Enc(-\tilde{\mathbf{H}})$)
7: **return** encryption $Enc(\mathbf{L})$
---

we omit the details for brevity. Conversion between Paillier encryption and garbled circuit follows popular approaches [Nikolaenko et al., 2013, Xie et al., 2014]). *Note that the whole phase only needs to occur once, which is a significant improvement over Newton method-based protocols.*

The second phase (Steps 2 to 14 in Algorithm 1) of PrivLogit-Hessian resembles that of the widely-used privacy-preserving distributed Newton method [Li et al., 2016], except for the substitution of repeated Hessian evaluation and inversion. Model estimation proceeds in a secure and iterative process (each execution within *while* loop constitutes an iteration, as shown in Algorithm 1). Model convergence is securely checked at each iteration using secure comparison (Steps 12 or 2 in Algorithm 1). For each iteration, local organizations only need to compute their local gradient $\mathbf{g}_j$ and log-likelihood $l_{sj}$ (where $j$ indexes each organization) using privacy-free computation, and securely transmit their (Paillier) encryptions of these summaries to the Center (Steps 3 to 7). The Center securely aggregates the gradient and log-likelihood submissions (using Paillier-based secure addition), and compose the encrypted global gradient (Step 8) and log-likelihood (Step 11). Later on in Step 9, the textbook method back-substitution (similar to [Nikolaenko et al., 2013]) is securely performed to derive the encrypted product $Enc(\tilde{\mathbf{H}}^{-1}\mathbf{g})$ from previously derived encryptions $Enc(\mathbf{L})$ and $Enc(\mathbf{g})$. The Center then updates current coefficient estimates following the PrivLogit updating formula (Step 10 and Equation 8). This iterative process continues until model converges.

## 4.2 PrivLogit-Local: Decentralizing More Computations.

Our second and even faster secure protocol, PrivLogit-Local, is presented in Algorithm 3. This protocol takes advantage of the fact that the centrally aggregated approximate Hessian $\tilde{\mathbf{H}}^{-1}$ (or encryption $Enc(\tilde{\mathbf{H}}^{-1})$) can be regarded as a (private) *constant* value. For each local Node $j$, local gradient $\mathbf{g}_j$ is privacy-free and essentially a *public constant*. This means that we can further distribute the expensive (Center-based) secure matrix-vector multiplication (Step 9 in Algorithm 1 of PrivLogit-Hessian) to local Nodes by leveraging much cheaper secure multiplication-by-constant operations locally: i.e., to locally compute multiplication $Enc(\tilde{\mathbf{H}}^{-1}) \otimes \mathbf{g}_j$ at each Node (which can be centrally aggregated efficiently in secure later) instead of at the Center.

In greater detail, the first step of PrivLogit-Local still involves the local organizations and Center securely approximating and "inverting" the Hessian (Step 1 in Algorithm 3; or *SetupOnce()* in Algorithm 2),

**Algorithm 3** PrivLogit-Local: offsetting partial Newton update step to local organizations.

**Input:** Random initial $\boldsymbol{\beta}^{(0)}$; regularization parameter $\lambda$
**Output:** Globally fit coefficient estimate $\boldsymbol{\beta}$

*[At local organizations and Center]* :

1: Securely approximate and Cholesky-decompose Hessian: $Enc(\mathbf{L}) = SetupOnce()$ (where $Enc(\mathbf{LL}^T) = Enc(-\tilde{\mathbf{H}})$) (Algorithm 2)
2: Securely invert Hessian: $Enc(\tilde{\mathbf{H}}^{-1}) \leftarrow Enc(\mathbf{L})$

3: **while** regression model not converged **do**
   *[At local organizations]* :
4:    **for** each organization j = 1 **to** S **do**
   *[At local Organization j]* :
5:       Compute local log-likelihood $l_{sj}$ and encrypt (Equation 9)
6:       Compute local gradient $\mathbf{g}_j$ (Equation 4)
7:       Secure multiplication-by-constant: $Enc(\tilde{\mathbf{H}}^{-1}\mathbf{g}_j) \leftarrow Enc(\tilde{\mathbf{H}}^{-1}), \mathbf{g}_j$;
8:       Securely send encryptions $Enc(\tilde{\mathbf{H}}^{-1}\mathbf{g}_j), Enc(l_{sj})$ to Center
9:    **end for**

   *[At Center]* :
10:    Securely compose global numerical updating step: $Enc(\tilde{\mathbf{H}}^{-1}\mathbf{g}) = Enc(\tilde{\mathbf{H}}^{-1}\mathbf{g}_1) \oplus ... \oplus Enc(\tilde{\mathbf{H}}^{-1}\mathbf{g}_j) \oplus ... \oplus Enc(\tilde{\mathbf{H}}^{-1}\mathbf{g}_S) \ominus Enc(\lambda\tilde{\mathbf{H}}^{-1}\boldsymbol{\beta}^{(t)})$
11:    Securely update coefficient estimates via PrivLogit: $\boldsymbol{\beta}^{(t+1)} \leftarrow \boldsymbol{\beta}^{(t)}$ (Equation 8)
12:    Securely aggregate log-likelihood across organizations: $Enc(l_2) = Enc(l_{s1}) \oplus Enc(l_{s2}) \oplus ... \oplus Enc(l_{sj}) \ominus Enc(\frac{\lambda}{2}[\boldsymbol{\beta}^{(t)}]^T\boldsymbol{\beta}^{(t)})$ (Equation 9)
13:    Securely check model convergence using secure comparison (Section 3.2)
14:    Securely disseminate new coefficient estimates to each local organiztions: $Enc(\boldsymbol{\beta}^{(t+1)})$
15: **end while**
16: **return** $\boldsymbol{\beta}^{(t)}$ (last converged estimate)

similar to Phase 1 of PrivLogit-Hessian. Next, we directly materialize the inversion of approximate Hessian in encrypted form, i.e., $Enc(\tilde{\mathbf{H}}^{-1})$. After that, this encrypted inversion is disseminated to each local organizations where local computation of gradients only involves privacy-free operations.

Later on, at each iteration, local organizations derive their various local summaries (just as the standard secure Newton method or our PrivLogit-Hessian), such as log-likelihood (Step 5) and gradient (Step 6). In addition, per earlier observation, they compute their respective versions of (partial) Newton updating step, by using efficient secure multiplication primitives (Step 7). Since the local gradients $\mathbf{g}_j$ do not involve privacy concerns at their respective local organizations (thus can be regarded as a public constant value), the computation is greatly simplified to highly efficient secure multiplication-by-constant primitives in Paillier. Afterwards, local organizations securely send their respective encrypted summaries $Enc(\tilde{\mathbf{H}}^{-1}\mathbf{g}_j), Enc(l_{sj})$ back to the Center (Step 8). For regularized logistic regression, the regularization term also needs to be securely composed, which can be prepared by the local organizations and then aggregated centrally, i.e., $Enc(\lambda\tilde{\mathbf{H}}^{-1}\boldsymbol{\beta}^{(t)}) = Enc(\tilde{\mathbf{H}}^{-1}\sum_{j=1}^{S}\lambda\boldsymbol{\beta}_j^{(t)})$. Finally, the Center only needs to perform trivial secure aggregation to complete the PrivLogit model updating process and secure convergence check (Steps 10 to 13).

The correctness of Algorithm 3 is straightforward (i.e., offsetting matrix multiplication to local Nodes). Briefly, $\tilde{\mathbf{H}}^{-1}\mathbf{g} = \tilde{\mathbf{H}}^{-1}(\sum_j \mathbf{g}_j - \lambda\boldsymbol{\beta}) = \sum_j \tilde{\mathbf{H}}^{-1}\mathbf{g}_j - \lambda\tilde{\mathbf{H}}^{-1}\boldsymbol{\beta}$.

Building on top of PrivLogit-Hessian, our second protocol PrivLogit-Local further avoids expensive secure matrix multiplication (between encryptions), which leads to *significantly simplified computation* and less overhead than PrivLogit-Hessian and baseline Newton. *Due to simplicity of computation equation, this also makes PrivLogit-Local more widely amenable to a variety of cryptographic schemes.*

## 5 Theoretical Proof and Analysis

In this section, we present theoretical analysis and proof for our proposals regarding computational complexity, and model convergence.

### 5.1 Proof on Model Convergence and Result Accuracy

Since our PrivLogit introduced approximation to Hessian, the convergence properties of standard Newton no longer apply. We thus present theoretical proof regarding the convergence properties of PrivLogit, which is based on quadratic function approximation [Böhning and Lindsay, 1988]. *We show that our PrivLogit optimizer is guaranteed to converge to the optimum (i.e., it always converges and converges with exact model accuracy), and at a linear convergence rate.* Specifically, we prove the following proposition:

**Proposition 1.** *Assume the optimal solution $\boldsymbol{\beta}^*$ to the objective function $l_2(\boldsymbol{\beta})$ (Equation 2) exists and is unique. Let $\{\boldsymbol{\beta}^{(t)}\}$ be a sequence generated by PrivLogit with the update formula in Equation 8. The sequence has the following properties:*

*(a) $l_2(\boldsymbol{\beta}^{(t+1)}) > l_2(\boldsymbol{\beta}^{(t)})$ and $\boldsymbol{\beta}^{(t)}$ will converge to the optimal solution $\boldsymbol{\beta}^*$.*

*(b) The rate of convergence of PrivLogit method is linear.*

*Proof.*
(a) By using the negative definiteness of $\tilde{\mathbf{H}}$ and the second-order Taylor expansion of $l_2(\boldsymbol{\beta})$, we have,

$$\begin{aligned}
&l_2(\boldsymbol{\beta}^{(t+1)}) - l_2(\boldsymbol{\beta}^{(t)}) \\
&= -\mathbf{g}(\boldsymbol{\beta}^{(t)})^\intercal \tilde{\mathbf{H}}^{-1}\mathbf{g}(\boldsymbol{\beta}^{(t)}) + \frac{1}{2}\mathbf{g}(\boldsymbol{\beta}^{(t)})^\intercal \tilde{\mathbf{H}}^{-1}\mathbf{H}(\hat{\boldsymbol{\beta}})\tilde{\mathbf{H}}^{-1}\mathbf{g}(\boldsymbol{\beta}^{(t)}) \\
&> -\mathbf{g}(\boldsymbol{\beta}^{(t)})^\intercal \tilde{\mathbf{H}}^{-1}\mathbf{g}(\boldsymbol{\beta}^{(t)}) + \frac{1}{2}\mathbf{g}(\boldsymbol{\beta}^{(t)})^\intercal \tilde{\mathbf{H}}^{-1}\tilde{\mathbf{H}}\tilde{\mathbf{H}}^{-1}\mathbf{g}(\boldsymbol{\beta}^{(t)}) \\
&= -\frac{1}{2}\mathbf{g}(\boldsymbol{\beta}^{(t)})^\intercal \tilde{\mathbf{H}}^{-1}\mathbf{g}(\boldsymbol{\beta}^{(t)}) > 0
\end{aligned}$$

where $\hat{\boldsymbol{\beta}}$ is between $\boldsymbol{\beta}^{(t)}$ and $\boldsymbol{\beta}^{(t+1)}$.

The objective function $l_2(\boldsymbol{\beta})$ is strictly concave with a negative definite Hessian matrix and therefore is maximized at the optimal solution $\boldsymbol{\beta}^*$. From the previous derivation, we obtain the lower bound of the increment of the objective function at each iteration. If $\mathbf{g}(\boldsymbol{\beta}^{(t)})$ is bounded away from 0 for all $t$, in other words, $||\mathbf{g}(\boldsymbol{\beta}^{(t)})|| > \epsilon$ for some positive constant $\epsilon$, then the increment of each iteration is also bounded above 0, which contradicts the upper boundedness of the objective function. Therefore, $\mathbf{g}(\boldsymbol{\beta}^{(t)}) \to 0$ as $t \to \infty$, which means the sequence $\{\boldsymbol{\beta}^{(t)}\}$ converges to the optimal solution $\boldsymbol{\beta}^*$.

(b) Since $\mathbf{X}^\intercal \mathbf{X}$ is positive semi-definite, its eigenvalues are all non-negative. Denote the biggest eigenvalue of $\mathbf{X}^\intercal \mathbf{X}$ as $\lambda_{max}$. Furthermore, we also assume $\mathbf{X}^\intercal \mathbf{A} \mathbf{X}$ is positive definite at every iteration, with the smallest eigenvalue $\lambda^{min} > 0$. Then we have

$$-\frac{1}{\frac{1}{4}\lambda_{max} + \lambda}\mathbf{I} \succeq \tilde{\mathbf{H}}^{-1}$$

and

$$-(\lambda^{min} + \lambda)\mathbf{I} \succeq \mathbf{H}(\boldsymbol{\beta}) \succeq -(\frac{1}{4}\lambda_{max} + \lambda)\mathbf{I}$$

Let $M = \frac{1}{4}\lambda_{max} + \lambda$ and $m = \lambda^{min} + \lambda$. By the strong concavity assumption and the second-order Taylor expansion of $l_2$, we have for any $\boldsymbol{v}$ and $\boldsymbol{\omega}$ in the parameter space,

$$l_2(\boldsymbol{\omega}) < l_2(\boldsymbol{v}) + \mathbf{g}(v)^\intercal (\omega - v) - \frac{1}{2}m\|\omega - v\|_2^2$$

$$< l_2(\boldsymbol{v}) + \frac{\|\mathbf{g}(\boldsymbol{v})\|_2^2}{2m}$$

Since the inequality holds everywhere in the parameter space, we have $\|g(\boldsymbol{v})\|_2^2 > 2m(l_2(\boldsymbol{\beta}^*) - l_2(\boldsymbol{v}))$ for any $\boldsymbol{v}$. Next we need to investigate the relation between $l_2(\boldsymbol{\beta}^*) - l_2(\boldsymbol{\beta}^{(t+1)})$ and $l_2(\boldsymbol{\beta}^*) - l_2(\boldsymbol{\beta}^{(t)})$ for all $t$. From part(a), we have

$$l_2(\boldsymbol{\beta}^{(t+1)}) > l_2(\boldsymbol{\beta}^{(t)}) - \frac{1}{2}\mathbf{g}(\boldsymbol{\beta}^{(t)})^\intercal \tilde{\mathbf{H}}^{-1} \mathbf{g}(\boldsymbol{\beta}^{(t)})$$

$$> l_2(\boldsymbol{\beta}^{(t)}) + \frac{1}{2M}\|\mathbf{g}(\boldsymbol{\beta}^{(t)})\|_2^2$$

Subtracting both sides from $l_2(\boldsymbol{\beta}^*)$, we get

$$l_2(\boldsymbol{\beta}^*) - l_2(\boldsymbol{\beta}^{(t+1)}) < l_2(\boldsymbol{\beta}^*) - l_2(\boldsymbol{\beta}^{(t)}) - \frac{1}{2M}\|\mathbf{g}(\boldsymbol{\beta}^{(t)})\|_2^2$$

$$< (1 - \frac{m}{M})(l_2(\boldsymbol{\beta}^*) - l_2(\boldsymbol{\beta}^{(t)}))$$

$$< (1 - \frac{m}{M})^t (l_2(\boldsymbol{\beta}^*) - l_2(\boldsymbol{\beta}^{(1)}))$$

where the factor $1 - \frac{m}{M} < 1$. It shows that $l_2(\boldsymbol{\beta}^{(t)})$ converges in a linear rate to $\boldsymbol{\beta}^*$ as $t \to \infty$. □

## 5.2 Complexity Analysis

Here we roughly analyze the computational complexity of the operations involved. Since cryptographic operations are dominating the total computation of secure protocols (often orders of magnitudes more expensive than non-secure computations), we thus focus on cryptography-related procedures only.

For gradient $\mathbf{g} \in \mathbb{R}^p$ and Hessian $\mathbf{H} \in \mathbb{R}^{p \times p}$, the main operations concerning cryptographic protection are: matrix inversion or closely related Cholesky decomposition (Equations 3 and 6) ($O(p^3)$ complexity), matrix-vector multiplication (Equations 3 and 6) ($O(p^2)$ complexity), and back-substitution (Step 9 in Algorithm 1) ($O(p^2)$).

State-of-the-art privacy-preserving Newton method requires repeated Hessian inversion and matrix multiplication, with total complexity of $O(p^3 \times \texttt{Newton iterations})$.

PrivLogit in general requires one step of Hessian inversion, and many iterations of matrix-vector multiplication, with total complexity of $O(p^3 + p^2 \times \texttt{PrivLogit iterations})$. Note that specifically for PrivLogit-Local, the second complexity term involving $p^2$ is much lower (i.e., much smaller constant factor in terms of empirical complexity) than PrivLogit-Hessian because secure multiplication-by-constant primitive (the main computation involved) is much more efficient than secure multiplication of two encryptions (as in PrivLogit-Hessian).

Since the strict relationship between $p$ and iteration numbers (of Newton and PrivLogit) is not determined, performance improvement is not strictly guaranteed for directly applying PrivLogit (as in the case of PrivLogit-Hessian) over Newton method. This is a limitation of one related alternative [Nardi et al., 2012]. In practice, we show that PrivLogit tends to have lower amortized cost, since PrivLogit iterations have very low cost. And this advantage grows with data dimensionality $p$. Our second adaption PrivLogit-Local is guaranteed to outperform Newton and the speedup is significant, because it replaces most iterations of Newton with extremely fast secure multiplication-by-constant steps.

## 5.3 Privacy Leaks and Security Guarantees

As is common for privacy-preserving distributed data mining [Aggarwal and Philip, 2008], our work follows the privacy definition of cryptography and secure multi-party computation, and assumes the *honest-but-curious* adversary model [Goldreich, 2001], where the adversary always follows the prescribed protocol, but may attempt to learn additional knowledge from the information flowed by. We emphasize that our secure protocols PrivLogit-Hessian and PrivLogit-Local directly adopt existing secure primitives and building blocks (i.e., garbled circuits [Yao, 1982], Paillier encryption [Paillier, 1999], and efficient conversion protocols), which are state-of-the-art and widely-used in the domain and whose security are well understood. Our work simply hybrids them in efficient ways to safeguard PrivLogit, whose security proof is straightforward based on security composition theorem [Goldreich, 2001]. Since detailed security analysis of such hybrid protocols is straightforward and widely available in secure Newton baseline and privacy-preserving (distributed) data mining [Nikolaenko et al., 2013, Aggarwal and Philip, 2008] and is not our focus or contribution (i.e., performance improvement from non-cryptography components), we only provide concise security analysis due to space constraint. Also, as is typical for cryptographic protocols [Aggarwal and Philip, 2008], our protocols only focus on protecting raw and intermediate data/computation, but not final result/output privacy. The latter belongs to a separate topic on differential privacy [Chaudhuri and Monteleoni, 2009] which is beyond the scope of our work.

Our protocols do disclose the regression coefficients (model output) to distributed Nodes and the number of iterations in numerical optimization. Most such leaks are inherent to the nature of cryptography in general (e.g., non-linear functions including logistic regression model cannot be directly and efficiently implemented fully in secure, requiring either revealing regression coefficients or noisy approximations), and numerical optimization (being iterative and distributed means that network communication pattern or iterations number is always observable). Since enhancing security of inherently hard problems is beyond the scope of this work and due to space constraint, our analysis only focuses on new materials not widely available in literature.

In addition, intermediate model outputs (revealed to local Nodes) share the same privacy properties as the final model output of the whole protocol, because they are all simply regression coefficients. By definition, cryptographic protocols do not protect final output, so this practice does not violate security. The only possible way to breach security in PrivLogit protocols is for local Nodes to accumulate all regression coefficients across all iterations and form a linear equation system to solve for individual record-level input values [O'Keefe and Chipperfield, 2013, El Emam et al., 2013]. However, since the number of iterations is quite small (at most linear in dimensionality $p$) and the (privacy-sensitive) data – the attack target – is huge in size ($n \times p$, where $n$ is extremely large), the system is severely underdetermined and such attacks are infeasible from information theory standpoint.

Except for the two aforementioned leaks, our secure protocols provides comprehensive security guarantees, leveraging proven cryptographic schemes and hybrid protocols. In both PrivLogit-Hessian and PrivLogit-Local, local-Node summaries are encrypted prior to submission to guarantee privacy. In PrivLogit-Local, the inverted approximate Hessian is also encrypted before being shared with local Nodes. At the aggregation Center, all incoming inputs are encrypted in Paillier or Yao's garbled circuit shares. All data, computations and results are also encrypted. Based on the composition theorem of security [Goldreich, 2001], the composition of these secure sub-protocols also yield a secure protocol overall.

# 6 Experiments

We implement both PrivLogit-Hessian and PrivLogit-Local in Java. Our garbled circuit is executed using state-of-the-art cryptography framework ObliVM-GC [Liu et al., 2015], which already provides thousands of times speedup over earlier works in our case. We use common privacy-preserving floating-point representations [Nikolaenko et al., 2013]. We use 2048-bit security parameter for encryption and other latest security parameters in ObliVM-GC. Secure Newton method and logistic regression has been explored by different communities and with different relaxations (to make computation feasible), but most were proposed prior to the fast ObliVM-GC framework and no open-source code is available as our baseline, making it difficult for direct and fair comparison. To set a directly comparable baseline, we thus also implement state-of-the-art privacy-preserving distributed Newton method using latest cryptography (same as our protocols), which may be of separate interest. We run all experiments between two commodity PCs with 2.5 GHz quad-core CPU and 16 GB memory, connected via ethernet.

Our empirical evaluations focus on the following criteria: 1) Model estimation quality (the accuracy of estimated coefficients) (in Section 6.2); 2) Model convergence performance (in Section 6.3). Due to space limit, we have to omit numerical stability results (that PrivLogit is always guaranteed to converge, while Newton is not), which has already been theoretically proved in Section 5.1 and does not affect our final conclusions.

In our experiments concerning numerical optimizers (i.e., PrivLogit and Newton method), we randomly initialize first coefficient estimates as commonly suggested (e.g., 0 as initial guess). We use $10^{-6}$ as our stopping threshold when checking model convergence (i.e., relative change of likelihood). Other thresholds have also been tested, such as $10^{-7}$ and $10^{-8}$, which does not affect our main conclusions and thus are omitted.

## 6.1 Datasets

Our empirical evaluation includes a series of simulated and real-world studies, covering a wide spectrum of applications from different domains and of different scales.

Among these, we have compiled four real-world studies, including: 1) the *Wine* quality study (with 6,497 samples and 12 features) [Cortez et al., 2009] for predicting wine quality from physicochemical tests, 2) online *Loans* data (with 122, 578 samples and 33 features) from Lending Club [LengingClub, 2016] for studying loan default status from loan application data, 3) company *Insurance* study (of dimension: $9,882 \times 38$) for predicting caravan issurance from demographic information and personal finance attributes, and 4) *News* dataset (of dimension $39,082 \times 52$) [Fernandes et al., 2015] for predicting the popularity of Mashable.com news from article features.

To make our evaluations more comprehensive, we have also simulated a series of studies with varying data scales, including: *SimuX10* (of dimension $50,000 \times 10$), *SimuX12* ($1,000,000 \times 12$), *SimuX50* ($1,000,000 \times 50$), *SimuX100* ($3,000,000 \times 100$), *SimuX150* ($4,000,000 \times 150$), *SimuX200* ($5,000,000 \times 200$), *SimuX400* ($50,000,000 \times 400$), etc. We also evaluated on additional studies with various data sizes and numbers of participating organizations. Since these factors do not have direct influence on the secure computation process (which primarily concerns summary data) both theoretically and empirically, we do not report on them separately. We following standard data simulation approach, by randomly generating covariates $\mathbf{X} \in R^{n \times p}$ and coefficients $\boldsymbol{\beta} \in R^{p \times 1}$, and then deriving responses $\mathbf{y} \in R^{n \times p}$ according to Bernoulli distribution.

These evaluation datasets should be representative for most large-scale studies in our focused domains in the foreable future. We also randomly partition datasets into subsets (by row or horizontally) in order to emulate different organizations in collaborative studies.

## 6.2 Model Accuracy

First, we demonstrate that our proposals obtain accurate results (i.e., regression coefficients). The standard non-secure distributed Newton method serves as the ground truth. Based on our theoretical proof on exact model accuracy (part of Section 5.1), our hypothesis is that despite the significant change in our numerical optimizer and reliance on cryptographic operations, the accuracy of our model estimation should still be guaranteed.

Numerical results have confirmed our hypothesis, as is illustrated in the QQ-plots in Figure 2. Specifically, our $\beta$ coefficient estimates from PrivLogit-Hessian and PrivLogit-Local are in perfect alignment with the ground-truth Newton across all studies, with correlation $R^2 = 1.00$ (perfect correlation and accuracy).

This implies that the approximate Hessian adaption we introduced in PrivLogit still maintains exact model accuracy. Moreover, it also confirms that the various cryptographic protections underlying PrivLogit-Hessian and PrivLogit-Local have no influence on the model quality.

## 6.3 Computational Performance

Next, we evaluate the computational performance of PrivLogit-Hessian and PrivLogit-Local in terms of model convergence with respect to iterations count and total runtime. We partition each evaluation datasets into 4∼20 blocks horizontally (i.e., by rows) to emulate different data-contributing organizations. As it has been demonstrated both theoretically and empirically that cryptographic protections do not affect the accuracy of computation in our case, we refer to our two secure protocols as PrivLogit in general for simplicity. Our model convergence threshold is set at $10^{-6}$, as mentioned earlier.

**Iterations to convergence** As is illustrated in Figure 3, all protocols managed to converge within a reasonable number of iterations. For instance, the *Loans* study (of dimension: $122,578 \times 33$) requires 6 and 17 iterations, respectively, to converge for the Newton and our PrivLogit-based secure proposals. For the smaller *Insurance* study, it takes 7 (for Newton) and 59 (for PrivLogit) iterations, respectively. As the data size (especially dimensionality) increases, we observe increases in the number of iterations both for Newton and PrivLogit, with the former growing slower. For instance, *SimuX150* (with 4 millions samples and 150 features) requires 7 iterations for Newton (only 17% increase over *Loans*) and 83 iterations for PrivLogit (388% increase over *Loans*).

Judging from model convergence iterations, PrivLogit seems "unfavorable" to Newton, as PrivLogit often requires a few tens of or more iterations, while the latter seems significantly faster with only single-digit number of iterations. The elongated convergence rate is perhaps the main reason why methods similar to PrivLogit have never been considered in the data security and privacy community. However, we will soon refute such a misconception by comparing the total runtime.

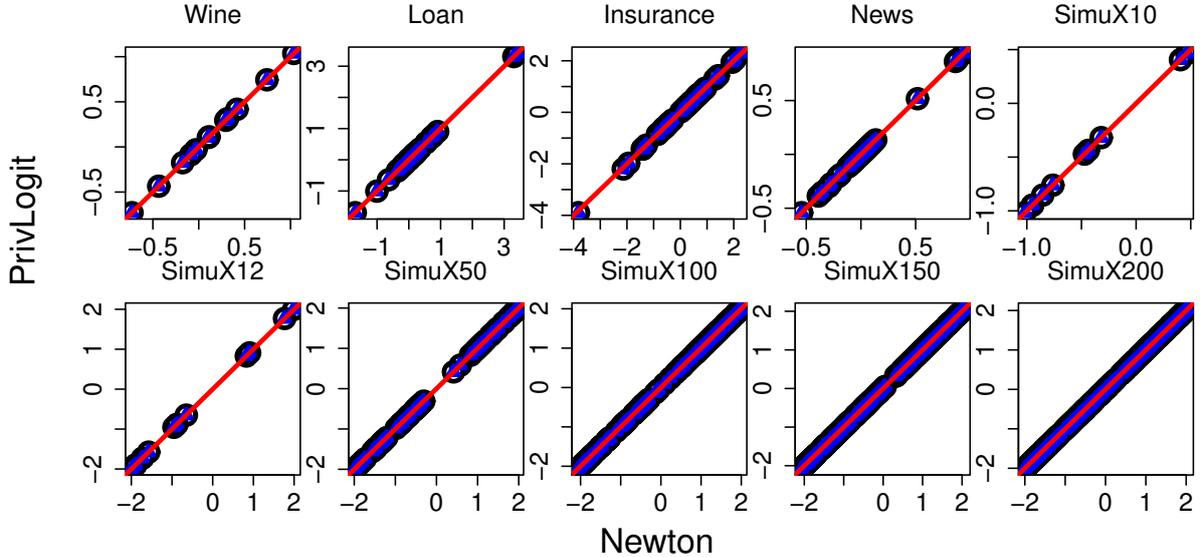

Figure 2: Result accuracy comparison across various datasets, as measured by QQ-plot of model coefficients estimated by ours (PrivLogit-Hessian and PrivLogit-Local; y-axis) vs. "ground-truth" Newton baseline (x-axis). PrivLogit-Hessian (in black) and PrivLogit-Local (in blue) points totally overlapped and both were perfectly aligned along the diagonal, showing exact result accuracy (perfect correlation of $R^2 = 1.00$ against baseline Newton).

**Convergence runtime** Surprisingly, detailed runtime benchmark in Table 2 manifests that both our secure protocols, i.e., PrivLogit-Hessian and PrivLogit-Local, turn out to be quite competitive in computational performance. For instance, in the *Loans* study, while Newton method takes only 6 iterations, its actual runtime reaches as much as 492 seconds (because of expensive per-iteration computation); On the other hand, despite requiring substantially more iterations (i.e., 17), our PrivLogit-Hessian and PrivLogit-Local protocols only take around 260 and 104 seconds, respectively, leading to 1.9x and 4.7x speedup, respectively. For even larger-scale studies such as *SimuX150*, Newton method takes 42,951 seconds or roughly 12 hours. PrivLogit-Hessian and PrivLogit-Local are respectively 1.7x and 7.1x times faster than Newton in this case.

One interesting observation is that in rare occasions, PrivLogit-Hessian can be slightly slower than Newton. For instance, the *Insurance* study requires around 843 seconds for Newton (for 7 iterations), but 978 seconds (1.16x slower) for PrivLogit-Hessian. This indicates that *directly* applying PrivLogit (i.e., PrivLogit-Hessian) does *not* guarantee improvement. Our second protocol, PrivLogit-Local, however, always outperforms Newton with dramatic speedup: requiring only 144 seconds (5.9x speedup).

Overall, PrivLogit-Local constantly outperforms other methods with significant speedup, while PrivLogit-Hessian is generally faster than Newton most of the time.

Furthermore, we also test on datasets with dimensions as high as 400, *a scale that has never been reported before for privacy-preserving logistic regression (not even for a much simpler linear regression*

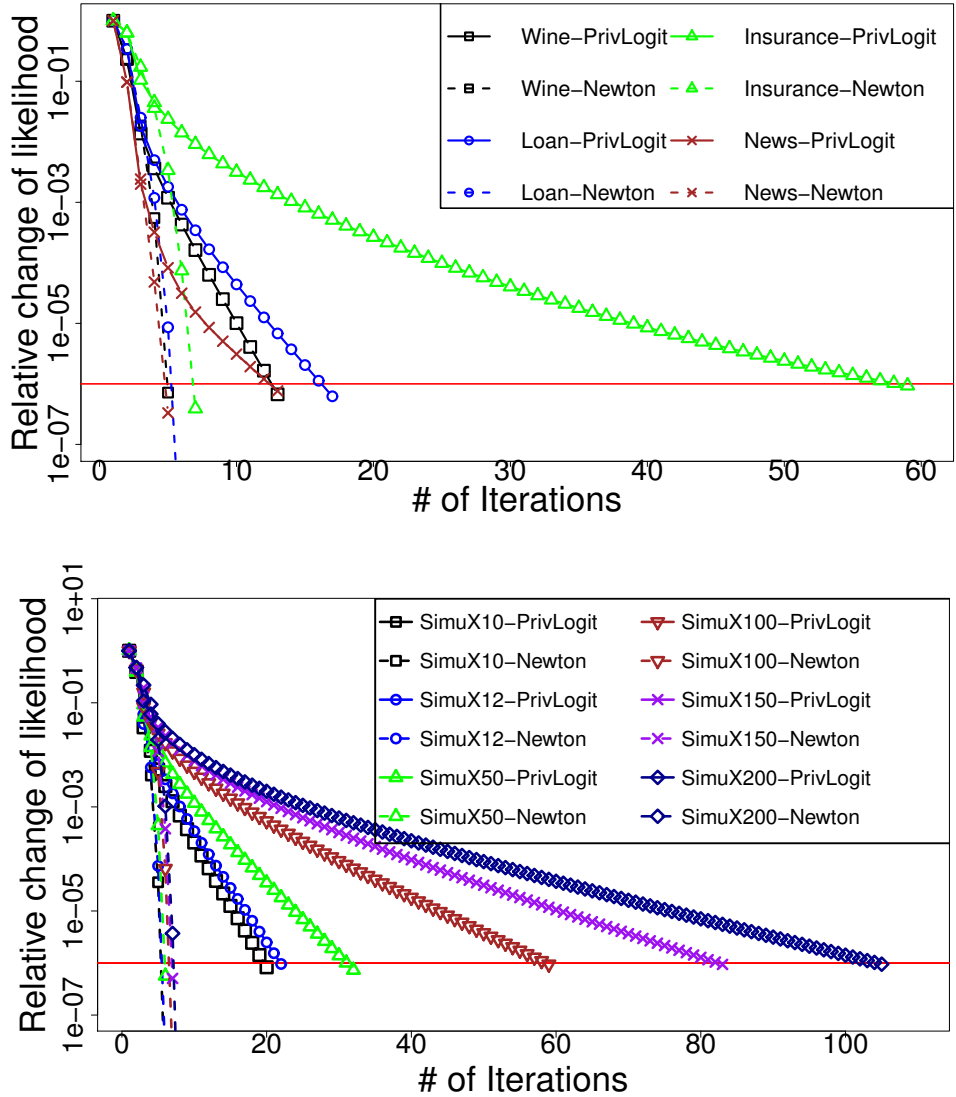

Figure 3: Convergence iterations of PrivLogit and the Newton method baseline on real-world (upper panel) and simulated (lower panel) datasets. Red horizontal line denotes the stopping threshold.

*model [Nikolaenko et al., 2013]).* Unfortunately, only PrivLogit-Local converges within reasonable time (110,598 seconds or roughly 1.28 days; for 206 iterations). The other two protocols still did not complete after 4 days. While PrivLogit-Hessian did not complete, its convergence iterations is expected to be the same as PrivLogit-Local (i.e., 206 iterations) since they implement the same optimization algorithm. For Newton method, a non-secure implementation requires 8 iterations.

**Relative speedup** To better demonstrate the relative performance of PrivLogit-Hessian and PrivLogit-Local over existing secure Newton methods, we extensively benchmark the relative speedup of our methods over the baseline Newton. As illustrated in Figure 4, PrivLogit-Hessian outperforms Newton most of the time (except for one occurrence of *Insurance*), and the speedup is between 1.03x∼2.32x. For PrivLogit-Local, the speedup is even more striking, with a speedup of up to 8.1x. For small datasets, PrivLogit-Local is around 2x faster than Newton; for medium datasets such as *Loans, Insurance, News*, its speedup is around 4x∼6x. The largest increase in relative performance is from PrivLogit-Local on the *SimuX200* dataset, with 8.1x speedup. PrivLogit-Hessian also performs well, with 2x speedup. In general, as data dimension increases, we see much more relative efficiency gain for both PrivLogit-Hessian and PrivLogit-Local.

Table 2: Model convergence iterations and runtime (in seconds) benchmark for Newton, PrivLogit, PrivLogit-Hessian, PrivLogit-Local.

| Dataset | Iterations Newton | Iterations PrivLogit | Time Newton | Time (S) PrivLogit-Hessian | Time (S) PrivLogit-Local |
|---|---|---|---|---|---|
| Wine | 5 | 13 | 32 | 24 | 17 |
| Loans | 6 | 17 | 492 | 260 | 104 |
| Insurance | 7 | 59 | 843 | 978 | 144 |
| News | 5 | 13 | 1442 | 621 | 313 |
| SimuX10 | 6 | 20 | 26 | 24 | 13 |
| SimuX12 | 6 | 22 | 38 | 37 | 17 |
| SimuX50 | 6 | 32 | 1549 | 1052 | 383 |
| SimuX100 | 7 | 59 | 13138 | 7817 | 1807 |
| SimuX150 | 7 | 83 | 42951 | 25030 | 6055 |
| SimuX200 | 8 | 105 | 114522 | 56917 | 14105 |
| SimuX400 | 8 | 206 | N/A | N/A | 110598 |

Overall, this provides further evidence that our secure PrivLogit proposals have better performance compared to state-of-the-art privacy-preserving distributed Newton method, and our relative competitive advantage increases along with data scale. This indicates that our methods hold much better potential for large-scale studies in the big data era.

## 7 Related Works

Privacy-preserving regression analysis and machine learning in general is actively investigated. Here we discuss several closely related lines of research.

### 7.1 Cryptographic Protections on Logistic Regression and Other Models

Extensive efforts have focused on protecting privacy in logistic regression, from centralized solutions [El Emam et al., 2013] to distributed architecture [Karr et al., 2007, Wolfson et al., 2010, Nardi et al., 2012, Li et al., 2015, Wu et al., 2012b,a]. Due to complexity of securely computing logistic function, many existing proposals compromise on security guarantee by providing no or only weak protections over intermediate summary data [Wolfson et al., 2010, Wu et al., 2012a, Li et al., 2016], which can be problematic given various inference attacks [O'Keefe and Chipperfield, 2013, Xie et al., 2014, El Emam et al., 2013]. Other works approximate the logistic function, resulting in accuracy loss [Nardi et al., 2012, El Emam et al., 2013, Aono et al., 2016]. Nearly all existing works directly apply mainstream model estimation algorithms (i.e., Newton method) without customization. Our proposal, however, provides a secure computing-centric perspective, and proposes a tailored optimizer specifically for cryptography that significantly outperforms alternatives while guaranteeing accuracy and privacy.

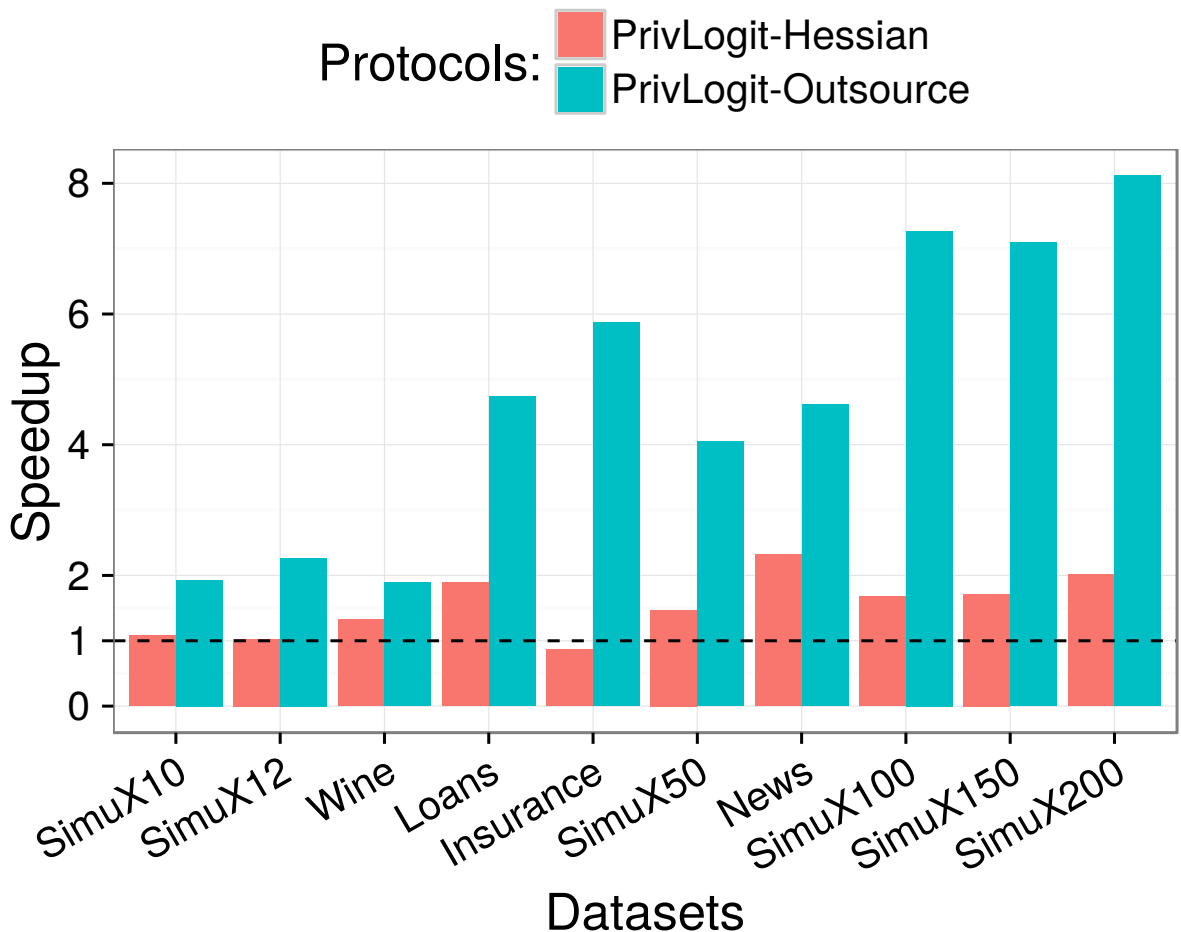

Figure 4: Relative speedup of PrivLogit-Hessian and PrivLogit-Local over the secure distributed Newton baseline (the $y = 1$ line), across various datasets. Our protocols can speed-up the computation by up to 2.32x and 8.1x (and even more), respectively.

Hessian approximation was briefly explored by [Nardi et al., 2012], but without justification or even (comparative) performance evaluation. Our results show that direct application of the method does not necessarily lead to better performance, and even when it does, the improvement is modest. In addition, for datasets of size $n \times p$, Newton method has per-iteration complexity $O(np^2 + p^3)$ (where the first term is dominating the cost). And the main improvement of approximate Hessian is by limiting the first term $np^2$ to one occurrence only (as in [Nardi et al., 2012]). However, our use case is different as our local-organization computation is privacy-free (i.e., independent from sample size $n$) and total cost is only determined by the second term $O(p^3)$, making it not obvious of the benefits of approximate Hessian. In fact, there is no performance guarantee if *directly* adopting approximate Hessian in our situation.

Cryptography is also widely used to safeguard other machine learning models, as partially reviewed by [Aggarwal and Philip, 2008].

### 7.2 Perturbation-based Privacy Protection

Perturbing data via artificial noise is also a popular technique for privacy preservation (e.g., $k-$anonymity, differential privacy [Chaudhuri and Monteleoni, 2009]). However, since such methods inherently change the data and computation output, their results may no longer be scientifically valid and thus are not widely accepted in practice. In addition, they do not protect the computation process, which is the central goal of cryptography-based protections.

### 7.3 Improved Numerical Optimization for Regression

Numerical optimization for regression analytics is under extensive investigation. These include various efforts to approximate or eliminate the Hessian from Newton-style optimizers, such as the Quasi-Newton or Hessian-free optimization (e.g., BFGS and L-BFGS [Liu and Nocedal, 1989]). However, none of them have seen adoption in data security and privacy research, partially because they are heavily tailored for privacy-free scenarios and often data-dependent and thus difficult for cryptographic implementation. Hessian approximation was described in the 1980s for maximum likelihood (in privacy-free applications) [Böhning and Lindsay, 1988], but only with limited adoption in practice maybe due to their not-obvious efficiency improvement for privacy-free settings.

## 8 Discussion & Conclusion

In PrivLogit-Hessian and PrivLogit-Local, the network bandwidth and transmission cost is small, since the encrypted summary information exchanged has very minimal size even for large studies, especially given that Hessian only needs to be preprocessed once. Since these factors are already accounted for in the total runtime benchmark, we omit detailed discussion for brevity.

The PrivLogit optimizer is designed for secure computing in general and agonistic of specific cryptographic schemes. Thus our empirical evaluation is focused on showing *further speedup on top of state-of-the-art cryptography*. There is room for further acceleration on our protocols as cryptography continues to improve, especially given that our computation is significantly simpler than baseline Newton. However, since our work focuses on *relative speedup* (excluding improvement in cryptography alone) and we aim to provide a direct comparison with state-of-the-art based on the same cryptography primitives, we leave it as future work to explore alternative cryptographic schemes.

While our work focuses on logistic regression model, our proposal of tailoring optimizers for secure computing seems widely applicable to privacy-preserving machine learning, as mainstream (distributed) numerical optimizers are not necessarily competitive for secure computing despite their wide and "direct" adoption in data security and privacy. We consider extending this novel approach to other machine learning models, such as other classifiers [Zheng et al., 2017], regressors, and deep learning [Beaulieu-Jones and Greene, 2016] that are increasingly popular in privacy-sensitive domains.

## 8.1 Conclusion

In this work, we have made a novel observation about a generic performance bottleneck in privacy-preserving logistic regression, and proposed an improved numerical optimizer (i.e., PrivLogit) and demonstrate its obscure but surprisingly competitive performance for privacy-preserving logistic regression. This contrasts to common practice in privacy-preserving data mining (i.e., directly applying mainstream numerical optimization), which often disregards secure computing-specific characteristics and thus misses valuable opportunities for significant performance boost. Based on PrivLogit, we also propose two secure and highly-efficient protocols for privacy-preserving logistic regression. We validate our proposals extensively using both analytical and empirical evaluations. Results indicate that our proposals outperform alternatives by several times while ensuring privacy and accuracy. Our methods should be helpful for making privacy-preserving logistic regression more scalable and practical for large collaborative studies. And our novel and generic perspective on tailoring optimizers for secure computing should also inspire other research in secure data management in general.